%% file: iclr_2026.tex
\PassOptionsToPackage{table}{xcolor}
\documentclass{article}

\usepackage{iclr2026_conference,times}
\iclrfinalcopy

\usepackage[utf8]{inputenc}
\usepackage[T1]{fontenc}
\usepackage{hyperref}
\usepackage{url}
\usepackage{graphicx}
\usepackage{booktabs}
\usepackage{multirow}
\usepackage{amsmath}
\usepackage{amsfonts}
\usepackage{amsthm}
\usepackage{algorithm}
\usepackage{algpseudocodex}
\usepackage{listings}
\usepackage[most]{tcolorbox}
\usepackage{paracol}
\usepackage{nicefrac}
\usepackage{microtype}
\usepackage{enumerate}
\usepackage{enumitem}

\definecolor{arxivblue}{HTML}{000080}

\hypersetup{
    colorlinks=true,
    linkcolor=arxivblue,
    citecolor=arxivblue,
    urlcolor=arxivblue
}

\input{chapter/def}

\lstdefinestyle{mocov3}{
  basicstyle=\ttfamily\footnotesize,
  breaklines=true,
  columns=fullflexible,
  keepspaces=true
}

\title{\methodname{}: Learning to Internalize Self-Critique \\with Reinforcement Learning}

\author{%
  \textbf{Jianbo Lin}$^{1,2}$\quad
  \textbf{Xiaomin Yu}$^{1}$\quad
  \textbf{Yi Xin}$^{2}$\quad
  \textbf{Yifu Guo}$^{3}$\quad
  \textbf{Zhuosong Jiang}$^{4}$\\
  \textbf{Zhongqi Yue}$^{7}$\quad
  \textbf{Weishi Wang}$^{6}$\quad
  \textbf{Heqing Zou}$^{5}$\quad
  \textbf{Chengwei Qin}$^{1\dagger}$\quad
  \textbf{Hui Xiong}$^{1}$\\[4pt]
  $^{1}$Hong Kong University of Science and Technology (Guangzhou)\quad
  $^{2}$Nanjing University\\
  $^{3}$Sun Yat-sen University\quad
  $^{4}$National University of Singapore\\
  $^{5}$Nanyang Technological University\quad
  $^{6}$SAP\quad
  $^{7}$Microsoft Research\\
  \texttt{jianbo.lin@outlook.com}
}

\begin{document}

\maketitle
\lhead{Preprint.}
\renewcommand{\thefootnote}{\fnsymbol{footnote}}
\footnotetext[2]{Corresponding author.}
\renewcommand{\thefootnote}{\arabic{footnote}}

\input{chapter/00_abstract}
\input{chapter/01_intro}
\input{chapter/02_preliminary}

\input{chapter/03_method}
\input{chapter/04_experiment}

\input{chapter/05_discussion}
\input{chapter/06_related_work}
\input{chapter/07_conclusion}

\medskip

{\small
\bibliographystyle{iclr2026_conference}
\bibliography{references}
}

\input{chapter/appendix}

\newpage

\end{document}

%% file: chapter/def.tex
\newcommand{\methodname}{ICRL}

%% file: chapter/00_abstract.tex
\begin{abstract}
Large language model-based agents make mistakes, yet critique can often guide the same model toward correct behavior. However, when critique is removed, the model may fail again on the same query, indicating that it has not internalized the critique's guidance into its underlying capability. Meanwhile, a frozen critic cannot improve its feedback quality over time, limiting the potential for iterative self-improvement. To address this, we propose learning to \textbf{I}nternalize self-\textbf{C}ritique with \textbf{R}einforcement \textbf{L}earning(\textbf{\methodname{}}), a novel framework that jointly trains a solver and a critic from a shared backbone to convert critique-induced success into unassisted solver ability. The critic is rewarded based on the solver's subsequent performance gain, incentivizing actionable feedback. To address the distribution shift between critique-conditioned and critique-free behavior, \methodname{} introduces a distribution-calibration re-weighting ratio that selectively transfers critique-guided improvements compatible with the solver's own prompt distribution. Additionally, a role-wise group advantage estimation stabilizes joint optimization across the two roles. Together, these mechanisms ensure that the solver learns to improve itself without external critique, rather than becoming dependent on critique-conditioned behavior. We evaluate \methodname{} on diverse benchmarks spanning agentic and mathematical reasoning tasks, using Qwen3-4B and Qwen3-8B as backbones. Results show consistent improvements, with average gains of 6.4 points over GRPO on agentic tasks, and 7.0 points on mathematical reasoning. Notably, the learned 8B critic is comparable to 32B critics while using substantially fewer tokens. The code is available at \href{https://github.com/brick-pid/ICRL}{\texttt{https://github.com/brick-pid/ICRL}}.

\end{abstract}

%% file: chapter/01_intro.tex
\section{Introduction}
\begin{figure*}[t]
    \centering
    \includegraphics[width=\textwidth]{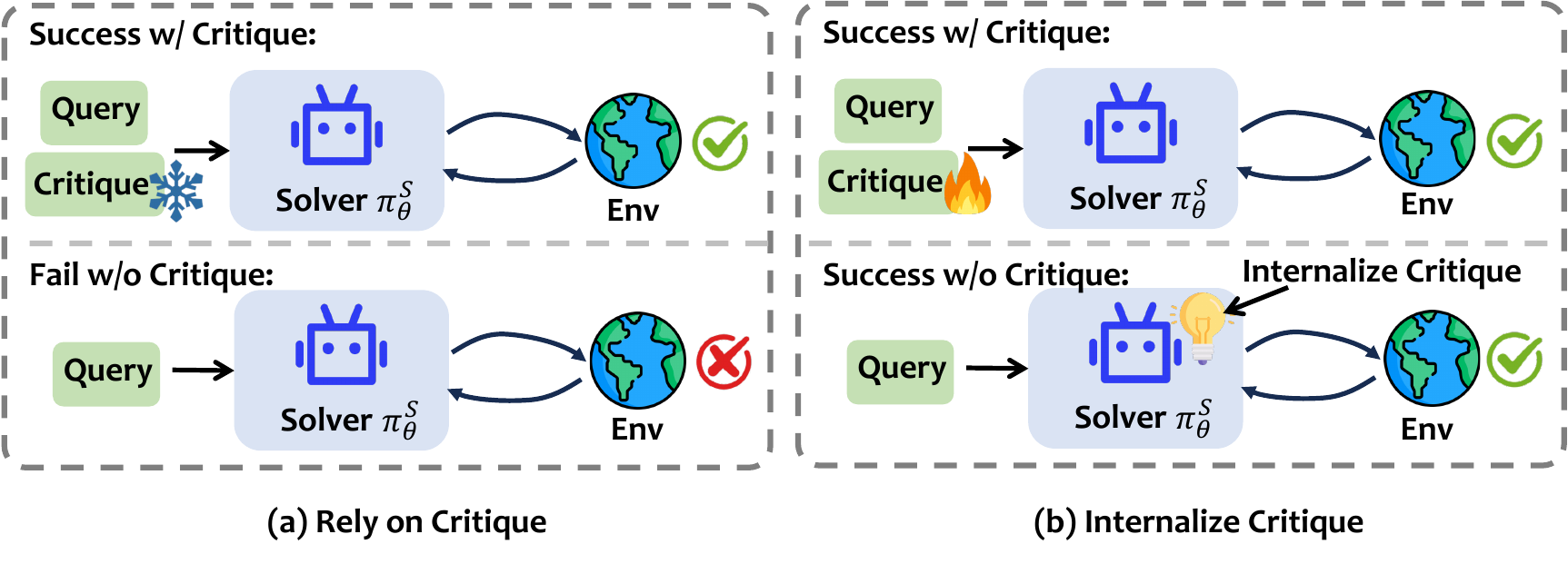}
    \vspace{-20pt}
    \caption{Critique can turn failed trajectories into successful revisions, while training should internalize such revision behavior into the critique-free solver.}
    \label{fig:intro}
    \vspace{-10pt}
\end{figure*}

Large language model (LLM)-based agents make mistakes when solving complex tasks \cite{yang2024sweagent,wang2024openhands,wang2024mobile,qin2025ui,li2025websailor,li2025chain}. As illustrated in Figure~\ref{fig:intro}, critique can guide the same model to correct its errors and successfully complete the task \cite{madaan2023self-refine,shinn2023reflexion,liu2025trust_verify,chen2026learning_self_verify,gou2023critic,Asai2023SelfRAGLT}. However, when critique is removed, the model may fail again on the same query, indicating that it has not internalized the guidance from critique into its underlying capability. Meanwhile, a frozen critic cannot improve its feedback quality over the course of training, limiting the potential for iterative self-improvement.

The challenge is \emph{internalization}: how can critique-guided revisions be transferred into the solver's critique-free policy? A fundamental distributional obstacle arises when critique-based self-improvement is introduced into training. As illustrated in Figure~\ref{fig:intro}(a), when an agent fails on query $q$ but succeeds after receiving critique $c$, the successful trajectory is sampled from a critique-conditioned behavior distribution $\pi(y \mid q, c)$ rather than the solver's original distribution $\pi(y \mid q)$. Training on such trajectories reinforces critique-dependent behavior, where the solver learns to perform well \emph{given} critique, not to perform well \emph{without} it \cite{scheurer2023training-feedback}. Without explicit correction for this distributional mismatch, standard policy optimization produces a biased estimate of the intended critique-free policy update. 

The quality of critique is equally important. The ability to diagnose the solver's failures and propose actionable corrections should itself be learnable and co-evolve with the solver. However, existing critique-based methods typically rely on a frozen critic model~\cite{zhang2025critique, tang2025self}, whose feedback quality remains static regardless of how the solver updates. This decoupling limits the agent's capacity for sustained self-improvement: as the solver advances, a stale critic may produce increasingly irrelevant or redundant feedback.

In this paper, we propose \textbf{\methodname{}}, a reinforcement-learning framework that converts critique-induced success into unassisted solver ability. \methodname{} jointly trains a solver and a critic from a shared backbone, without external teacher models or manually annotated critique data. The critic is rewarded for producing a critique that improves the solver's subsequent attempt, creating learning signals tied to direct critique utility. To internalize critique-guided revisions, \methodname{} reconditions revised trajectories under the solver's critique-free prompt and applies a token-level distribution-calibration re-weighting ratio. This ratio selectively transfers tokens whose generation is already plausible under the critique-free distribution, while down-weighting tokens that depend heavily on the critique context. The result is that the solver internalizes revision patterns compatible with its own prompt distribution. To further stabilize joint optimization, \methodname{} employs role-wise group advantage estimation that normalizes solver and critic rewards separately, preserving distinct learning signals for each role.

To our knowledge, \methodname{} is the first framework to improve both critique internalization and critic learning in a reinforcement-learning setting.
We evaluate \methodname{} on diverse environments, including text-world tasks \cite{ALFWorld20}, e-commerce web navigation \cite{yao2022webshop}, multi-hop question answering \cite{HotpotQA,2WikiMultiHopQA,Bamboogle,Musique}, and mathematical reasoning \cite{hendrycks2021math500,lewkowycz2022minerva,he-etal-2024-olympiadbench,numina_math_datasets}. Experiments are conducted on Qwen3-4B and Qwen3-8B, and compared with prompting-based baselines, reinforcement-learning baselines, and critique-based methods. Experimental results demonstrate empirical improvements over baselines on both agentic tasks and mathematical reasoning tasks.
We summarize our contributions as follows:
\begin{itemize}[leftmargin=*]
    \item We propose the \methodname{} framework, a solver-critic reinforcement-learning framework that enables iterative self-improvement by jointly learning to critique and to internalize critique.
    \item We introduce a distribution-calibration re-weighting ratio that corrects the distributional shift between critique-conditioned and critique-free behavior. We further propose role-wise group advantage estimation to stabilize joint solver-critic optimization.
    \item Experimental results demonstrate empirical improvements over baselines on both agentic and mathematical reasoning tasks. The jointly learned critic achieves performance comparable to 32B frozen critics while using fewer tokens.
\end{itemize}

%% file: chapter/02_preliminary.tex
\section{Preliminary} \label{sec:preliminary}

\subsection{Task Formulation}

Given a query $q \in \mathcal{Q}$, an LLM-based agent $\pi_\theta$, parameterized by $\theta$, interacts with the environment to sample a trajectory $\tau$, and receives reward $r(\tau)$ from a task evaluator. We model agentic tasks as a Partially Observable Markov Decision Process (POMDP) following \cite{xi2025agentgym, xi2025agentgym-rl}. At each time step $t$, the agent conditions on history $h_t = (q, o_0, a_0, \ldots, o_t)$ consisting of the initial query and the sequence of past observations and actions up to time $t$. Based on this history, the agent samples an action $a_t \sim \pi_\theta(\cdot \mid h_t)$. This interaction proceeds for $H$ steps, resulting in a complete trajectory $\tau = (q, o_0, a_0, \ldots, a_{H-1}, o_H)$. The training objective is
\begin{equation}
    J(\theta) = \mathbb{E}_{\tau \sim \pi_\theta}\left[r(\tau)\right].
\end{equation}

\subsection{Group-Relative Policy Optimization}

We adopt GRPO \cite{guo2025deepseek} as the underlying reinforcement-learning
primitive. For a query $q$, GRPO samples a group of candidate trajectories $\mathcal{G}(q) = \{\tau_1, \dots, \tau_G\}$ and computes
the group-normalized advantage as
$\hat{A}_i = \frac{r(\tau_i) - \operatorname{mean}_{j}\,r(\tau_j)}{\operatorname{std}_{j}\,r(\tau_j) + \delta}$,
where $\delta > 0$ is a small constant for numerical stability.
Let $y_t$ denote the $t$-th generated token of sample $\tau_i$; the
importance-sampling ratio
$\rho_t(\theta) = \frac{\pi_\theta(y_t \mid q, y_{<t})}{\pi_{\theta_{\mathrm{old}}}(y_t \mid q, y_{<t})}$
measures the deviation of the current policy from the behavior policy used to
collect the group. The resulting clipped surrogate objective is
\begin{equation}
  J_{\mathrm{GRPO}}(\theta)
  = \mathbb{E}_{i,\, t}
    \!\left[
      \min\!\Big(
        \rho_t(\theta)\,\hat{A}_i,\;
        \operatorname{clip}\!\big(\rho_t(\theta),\,1-\epsilon,\,1+\epsilon\big)
        \,\hat{A}_i
      \Big)
    \right],
\end{equation}
where $i$ indexes a trajectory in the sampled group $\mathcal{G}(q)$, $t$ indexes a generated token position within that trajectory, and $\epsilon$ is the clipping parameter.

%% file: chapter/03_method.tex
\section{Methodology}

\begin{figure*}[t]
    \centering
    \includegraphics[width=1.0\textwidth]{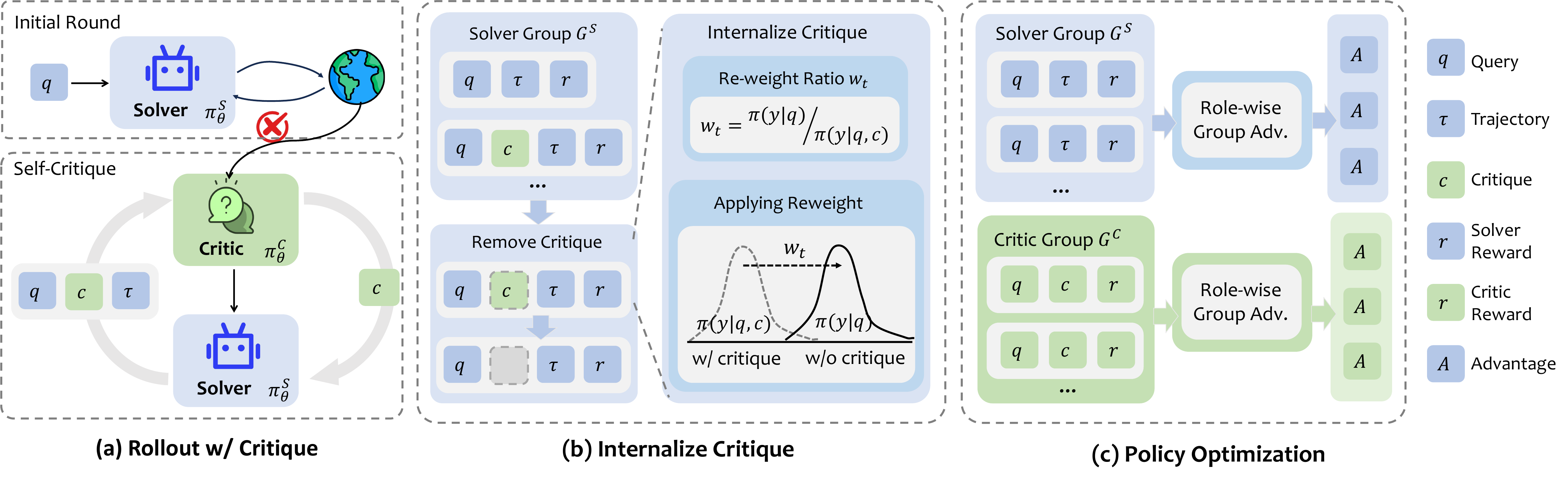}
    \vspace{-15pt}
    \caption{Overview of the \methodname{} framework.
    (1) Rollout with critique alternates solver and critic;
    (2) Policy optimization jointly trains both solver and critic;
    (3) Internalizing critique into solver.}
    \label{fig:overview}
    \vspace{-10pt}
\end{figure*}

\subsection{Self-Improving Workflow}
\label{sec:mas}

\textbf{Self-Improving Workflow.}
\methodname{} instantiates two agent roles, a solver and a critic, from a shared backbone with parameters $\theta$ and role-specific prompts $p^{\mathcal{S}}$ and $p^{\mathcal{C}}$.
Specifically, the \textit{solver} $\pi^\mathcal{S}_{\theta}$ generates task-solving trajectories, while the \textit{critic} $\pi^\mathcal{C}_{\theta}$ generates natural-language critique after failed attempts.
For each query $q$, we execute an iterative self-improvement session with at most $K$ total rounds.
The solver first samples an initial trajectory $\tau_1 \sim \pi^\mathcal{S}_\theta(\cdot \mid q)$, and the environment provides an outcome reward $r(\tau_1)$.
If the current trajectory $\tau_i$ fails, the critic analyzes it and generates critique $c_i \sim \pi^\mathcal{C}_\theta(\cdot \mid q,\tau_i)$.
The solver then generates a revised trajectory conditioned on the critique, $\tau_{i+1} \sim \pi^\mathcal{S}_\theta(\cdot \mid q,c_i)$.
This iterative process repeats until the task succeeds or the round budget is exhausted.
A complete rollout session may thus contain multiple rounds of solver trajectories and critic outputs, denoted as $S = (\tau_1, c_1, \tau_2, \ldots, c_{k-1}, \tau_k)$ with $k \leq K$.

\textbf{Solver and Critic Reward.}
The policy optimization is guided by separate reward signals for each role. For the solver, we use the task outcome reward $r(\tau)\in [0,1]$ for optimization, where $r(\tau)=1$ indicates successful task completion. For the critic, the reward measures whether its critique improves the solver's subsequent performance.
Specifically, after the critic produces a critique $c_i$ for the solver trajectory $\tau_i$ ($i < k$), the solver generates a revised trajectory $\tau_{i+1}$ conditioned on $c_i$.
As shown in Eq.~\eqref{eq:critic-reward}, the critic receives a reward of $1$ if the revised trajectory succeeds; otherwise, its reward equals the temporal improvement in the solver's reward, which is nonzero only when the environment provides non-binary dense rewards.

\begin{equation}
    \label{eq:critic-reward}
    r(c_i) = 
    \begin{cases}
        1, & \text{if } \tau_{i+1} \text{ succeeds}, \\
        r(\tau_{i+1}) - r(\tau_i), & \text{otherwise}.
    \end{cases}
\end{equation}
This design provides learning signals tied to downstream revision utility, where the critic is rewarded not for producing plausible-sounding feedback, but for feedback that demonstrably helps the solver.

\subsection{Self-Improvement Policy Optimization}

Self-improvement introduces two challenges absent from standard policy optimization. \textbf{(i) Mixed prompt prefixes.} GRPO relies on group-relative comparisons among samples generated under a shared prompt prefix and behavior distribution. In our workflow, however, initial attempts, critiques, and revisions are generated under different role-specific and information-conditioned prompts, so their rewards are not directly comparable within a single group. \textbf{(ii) Distributional shift.} The critic changes the solver's effective sampling distribution. Initial attempts are drawn from the critique-free solver behavior distribution $\pi_{\theta_{\mathrm{rollout}}}^{\mathcal S}(\cdot\mid q)$, whereas revised attempts are drawn from a critique-conditioned behavior distribution $\pi_{\theta_{\mathrm{rollout}}}^{\mathcal S}(\cdot\mid q,c)$. Treating revised trajectories as if they were sampled from the critique-free solver distribution yields a biased estimate of the intended policy update (see Appendix~\ref{appendix:dist-shift} for detailed analysis).

The revised trajectories are valuable because they contain solutions discovered through the critic's feedback. However, these trajectories are sampled from a critique-conditioned behavior distribution $\pi_{\theta_{\mathrm{rollout}}}^{\mathcal S}(\cdot\mid q,c)$, they cannot be used to update the critique-free policy without correction.

\textbf{Critique-Conditioned Distribution Calibration.}
To optimize critique-guided revisions under the solver's original behavior distribution, we recondition each revised trajectory by removing the critique from its prompt and treating it as evidence for $\pi_{\theta}^{\mathcal S}(\cdot\mid q)$.
We then introduce a token-level reweight ratio to calibrate this behavior distribution mismatch. For a self-improved round $(\tau_i, c_i, \tau_{i+1})$, where $\tau_i\sim \pi_{\theta_{\mathrm{rollout}}}^{\mathcal S}(\cdot\mid q)$ and $\tau_{i+1}\sim \pi_{\theta_{\mathrm{rollout}}}^{\mathcal S}(\cdot\mid q,c_i)$. We define the token-wise reweight ratio as

\begin{equation}
w_t
=
\begin{cases}
\frac{\pi^\mathcal{S}_{\theta_{\mathrm{rollout}}}(y_t \mid q, y_{<t})}{\pi^\mathcal{S}_{\theta_{\mathrm{rollout}}}(y_t \mid q, c, y_{<t})}
\,, & \text{critique-guided solver trajectories}, \\
1, & \text{otherwise}.
\end{cases}
\label{eq:reweight-ratio}
\end{equation}

\input{chapter/algorithm}

This ratio measures whether a critique-guided token is already plausible under the critique-free solver behavior distribution. Tokens whose probability remains high without the critique are transferred strongly to $\pi^\mathcal{S}_{\theta}(\cdot \mid q)$, while tokens that depend heavily on the critic context are downweighted. When $w_t \approx 1$, the token's generation does not depend on the critique context and can be directly transferred to the critique-free policy. When $w_t \ll 1$, 
the token relies heavily on the critique and is downweighted to avoid reinforcing critique-dependent behavior. Conversely, $w_t > 1$ indicates tokens that the critique-free solver would have generated with higher probability, which are upweighted. In this way, the solver internalizes revision behavior that is compatible with its own prompt distribution instead of blindly imitating critic-assisted outputs.

Following the GRPO importance-sampling ratio introduced in Section~\ref{sec:preliminary}, we use $\rho_t(\theta)$ to denote the token-level importance-sampling ratio.
For solver trajectories, this ratio is computed under the critique-free prompt context $(q,y_{<t})$ after removing the critique; critic trajectories are evaluated under their original prompt $(q,\tau_i)$.
The reweight ratio $w_t$ defaults to $1$ for all non-revised trajectories, while $\rho_t(\theta)$ retains its standard role as the clipped GRPO importance-sampling ratio.

\textbf{Role-wise Advantage Estimation.}
Solver and critic trajectories are generated under different prompt prefixes with distinct reward functions: the solver is optimized for task completion, while the critic is optimized for revision utility. Directly normalizing these heterogeneous rewards within a single group would violate the relative-comparison principle and produce misleading advantage estimates. We therefore compute the group-relative advantage separately for each role. Formally, for each query $q$ and role $g\in\{\mathcal{S},\mathcal{C}\}$, we collect a role-specific group $\mathcal{G}^{g}(q)=\{\tau_1^g,\dots,\tau_{G_g}^g\}$ and compute the advantage within that group:

\begin{equation}
\label{eq:subgroup-adv}
\hat{A}_i^{g}
=
\frac{
    r(\tau_i^g)-\operatorname{mean}_{j}\,r(\tau_j^g)
}{
    \operatorname{std}_{j}\,r(\tau_j^g)+\delta
},
\quad
g\in \{\mathcal{S}, \mathcal{C}\}.
\end{equation}

where $\delta>0$ is a small constant for numerical stability. This role-wise baseline preserves the relative-comparison principle while respecting the distinct semantics of each role.

\textbf{Policy Optimization Objective.}
Our final objective is a GRPO-style clipped update over sampled trajectories and token positions.
For brevity, we omit the role superscript in $\rho_t(\theta)$ when the role is clear from the trajectory $\tau$:
\begin{equation}
\label{eq:mixed-objective}
J(\theta)
=
\mathbb{E}_{\tau,\, t}
\!\left[
\min\!\big(w_t, w_\mathrm{max}\big)\,
\min\!\Big(
\rho_t(\theta)\hat{A}(\tau),\;
\operatorname{clip}\!\big(\rho_t(\theta),1-\epsilon,1+\epsilon\big)\hat{A}(\tau)
\Big)
\right],
\end{equation}

where only critique-guided revised solver trajectories receive the distribution-calibration reweight ($w_t$ from Eq.~\eqref{eq:reweight-ratio}), while initial solver and critic trajectories retain $w_t=1$.
The upper bound $w_\mathrm{max}$ prevents excessively large weights that could arise when the critique-free probability substantially exceeds the critique-conditioned probability, thereby bounding gradient variance. Since the solver and critic are two prompted roles instantiated from the same backbone, maximizing Eq.~\eqref{eq:mixed-objective} jointly trains both roles through their role-specific advantages and rewards.
The solver learns from initial and calibrated revised trajectories, progressively internalizing critique-guided improvements into its critique-free policy. The critic simultaneously learns to produce feedback that maximizes the improvement of the downstream solver.

%% file: chapter/algorithm.tex
\begin{algorithm}[t]
    \caption{ICRL Policy Optimization}
    \label{alg:mixed-policy-training}
    \begin{algorithmic}[1]
        \Require Problem batch $\mathcal{Q}$, group size $G$, maximum rounds $K$, current parameters $\theta$
        \For{each problem $q \in \mathcal{Q}$}
            \State $\mathcal{G}^{\mathcal{S}}(q) \gets \emptyset$, $\mathcal{G}^{\mathcal{C}}(q) \gets \emptyset$
            \For{$G$ independent sessions}
                \State Sample $\tau_1 \sim \pi^\mathcal{S}_{\theta_{\mathrm{rollout}}}(\cdot \mid q)$
                \State Compute solver reward $r^{\mathcal{S}}(\tau_1) \in [0,1]$
                \State Add $(q, \tau_1, r^{\mathcal{S}}(\tau_1))$ to $\mathcal{G}^{\mathcal{S}}(q)$
                \For{$i=1$ to $K-1$}
                    \If{$r^{\mathcal{S}}(\tau_i) = 1$}
                        \State \textbf{break}
                    \EndIf
                    \State Sample $c_i \sim \pi^\mathcal{C}_{\theta_{\mathrm{rollout}}}(\cdot \mid q, \tau_i)$
                    \State Sample $\tau_{i+1} \sim \pi^\mathcal{S}_{\theta_{\mathrm{rollout}}}(\cdot \mid q, c_i)$
                    \State Compute solver reward $r^{\mathcal{S}}(\tau_{i+1})$
                    \State Compute critic reward $r^{\mathcal{C}}(c_i)$ via Eq.~\eqref{eq:critic-reward}
                    \State Add $(q, \tau_i, c_i, r^{\mathcal{C}}(c_i))$ to $\mathcal{G}^{\mathcal{C}}(q)$
                    \State Add $(q, c_i, \tau_{i+1}, r^{\mathcal{S}}(\tau_{i+1}))$ to $\mathcal{G}^{\mathcal{S}}(q)$
                \EndFor
            \EndFor
            \State Compute $\hat{A}_i^{\mathcal{S}}$ and $\hat{A}_i^{\mathcal{C}}$ via Eq.~\eqref{eq:subgroup-adv}
            \State Compute reweight ratios $w_t$ for revised solver trajectories via Eq.~\eqref{eq:reweight-ratio}
            \State Update $\theta$ by maximizing the multi-role objective $J(\theta)$ in Eq.~\eqref{eq:mixed-objective}
        \EndFor
    \end{algorithmic}
\end{algorithm}

%% file: chapter/04_experiment.tex
\section{Experiments} \label{sec:experiments}

\subsection{Experimental Setup}
\textbf{Environments.} To comprehensively evaluate \methodname{}, we conduct experiments on four types of tasks.
(1) Text world: we use ALFWorld \cite{ALFWorld20}, a text-based environment that simulates embodied household tasks requiring multi-step navigation.
(2) Web navigation: we employ WebShop \cite{yao2022webshop}, an e-commerce website environment that requires agents to navigate, search for, and purchase products. 
(3) Multi-hop question answering: we evaluate on multi-hop question answering tasks using an RAG-based search environment, including HotpotQA \cite{HotpotQA}, 2WikiMultiHopQA \cite{2WikiMultiHopQA}, Bamboogle \cite{Bamboogle}, and MuSiQue \cite{Musique}. 
(4) Mathematical reasoning: we evaluate on five benchmarks, including MATH500 \cite{hendrycks2021math500}, Minerva Math \cite{lewkowycz2022minerva}, OlympiadBench \cite{he-etal-2024-olympiadbench}, AIME24 \cite{numina_math_datasets}, and AMC23 \cite{numina_math_datasets}. These datasets consist of high-school and college-level math problems. Detailed environment descriptions are provided in Appendix~\ref{sec:env}.

\textbf{Baselines and Backbone Models.} We compare our method with different baselines. For prompting-based baselines, we use off-the-shelf models, including Qwen3-4B\cite{yang2025qwen3}, Qwen3-8B\cite{yang2025qwen3}, Qwen3-30B-A3B\cite{yang2025qwen3}, Gemini-2.5-Flash\cite{comanici2025gemini25pushingfrontier}, and Gemini-3-Flash\cite{google2026gemini3flash}. For single-agent RL baselines, we include GRPO\cite{guo2025deepseek} and GSPO\cite{zheng2025group}. To compare against agent-oriented training methods, we further include ScalingInter-RL\cite{xi2025agentgym-rl}, which gradually increases the interaction horizon; MATPO\cite{mo2025matpo}, which trains the planner and subagent through role-specific policy optimization; and Critique-GRPO\cite{zhang2025critique}(self-critique), which introduces natural-language critiques for critique-guided policy optimization.
For backbone models, we consider Qwen3-4B and Qwen3-8B. 

\input{tables/01_exp.tex}

\subsection{Results on Agentic Tasks}

As shown in Table~\ref{tab:my_results}, \methodname{} improves agentic-task performance across the three agentic environments. On Qwen3-4B, \methodname{} achieves the best average score, reaching 57.0\%. It improves over GRPO by 7.8 points and surpasses Critique-GRPO by 1.1 points on average. Specifically, \methodname{} achieves the best success rate on ALFWorld and WebShop, indicating improved decision quality in long-horizon environments. On 2WikiMultiHopQA and MuSiQue, \methodname{} also achieves the best results, and on HotpotQA and Bamboogle, it remains competitive with the best-performing baselines. 

On Qwen3-8B, \methodname{} further obtains the highest average score of 57.8\%, improving over GRPO and Critique-GRPO by 5.0 points and 1.2 points, respectively. On the WebShop environment, \methodname{} achieves the best success rate and reward. On ALFWorld, \methodname{} also achieves competitive results. For multi-hop search tasks, \methodname{} achieves the best results on 2WikiMultiHopQA under both backbones and remains competitive on MuSiQue, HotpotQA, and Bamboogle, although it is not uniformly the best on all datasets. Overall, these results suggest that \methodname{} is effective on the evaluated agentic environments and backbone scales.

\subsection{Results on Math Tasks}

As shown in Table~\ref{tab:math}, \methodname{} performs effectively on mathematical reasoning tasks. On Qwen3-8B, SFT provides only moderate improvements, increasing the average score from 55.0\% to 59.2\%, while GRPO reaches 68.3\% by directly optimizing answer correctness. Critique-GRPO further improves the average score to 73.3\%, showing the benefit of critique information. \methodname{} achieves the best overall performance with an average score of 75.3\%, outperforming GRPO and Critique-GRPO by 7.0 points and 2.0 points, respectively. The improvements are especially clear on challenging competition-style benchmarks, such as OlympiadBench and AIME24, where \methodname{} improves over GRPO from 65.6\% to 68.9\% and from 50.0\% to 65.1\%. Compared with Critique-GRPO, \methodname{} performs better on four out of five benchmarks, with the only exception being AMC23. 
These math results are consistent with \methodname{} better internalizing critique-guided improvements, which is associated with stronger direct reasoning performance even without critique conditioning.

\input{tables/02_math.tex}

%% file: tables/01_exp.tex
\begin{table*}[!t]
\centering
\caption{Main results on agentic tasks. \textbf{Bold} indicates the best performance within each group.}
\label{tab:my_results}
\vspace{8pt} 
\resizebox{\linewidth}{!}{
\begin{tabular}{l cc c cccc c}
\toprule
\multirow{2.5}{*}{\textbf{Method}} & \textbf{ALFWorld} & \multicolumn{2}{c}{\textbf{WebShop}} & \textbf{HotpotQA} & \textbf{2Wiki} & \textbf{Bamboogle} & \textbf{MuSiQue} & \multirow{2.5}{*}{\textbf{Avg.}} \\
\cmidrule(lr){2-2} \cmidrule(lr){3-4} \cmidrule(lr){5-5} \cmidrule(lr){6-6} \cmidrule(lr){7-7} \cmidrule(lr){8-8}
& \textbf{SR} & \textbf{SR} & \textbf{Reward} & \textbf{SR} & \textbf{SR} & \textbf{SR} & \textbf{SR} &\\
\midrule
\multicolumn{9}{c}{\textit{Prompting}} \\
\midrule
Qwen3-4B & 22.0 & 1.5 & 9.6 & 24.0 & 32.0 & 26.0 & 6.0 & 17.3 \\
Qwen3-8B & 38.0 & 2.0 & 10.3 & 32.0 & 32.0 & 30.0 & 6.0 & 21.5 \\
Qwen3-30B-A3B & 45.5 & 3.0 & 25.6 & 40.0 & 58.0 & 44.0 & 16.0 & 33.2 \\
Gemini-2.5-flash & 78.5 & 41.2 & 48.7 & 47.0 & 66.0 & 59.5 & 26.0 & 52.4 \\
Gemini-3-Flash & 97.0 & 53.0 & 57.3 & 52.0 & 78.0 & 72.0 & 31.0 & 62.9 \\
\midrule
\multicolumn{9}{c}{\textit{Qwen3-4B}} \\
\midrule
GRPO & 81.0 & 63.5 & 81.8 & 38.0 & 36.0 & 30.0 & 14.0 & 49.2 \\
GSPO & 82.0 & 65.0 & 82.1 & 42.0 & 36.0 & 34.0 & 10.0 & 50.2 \\
ScalingInter & 90.0 & 66.5 & 84.7 & 44.0 & 38.0 & 32.0 & 15.0 & 52.9 \\
MATPO & 87.5 & 65.0 & 82.4 & \textbf{46.0} & 42.0 & \textbf{40.0} & \textbf{16.0} & 54.1 \\
Critique-GRPO & 91.5 & 72.5 & 87.1 & 43.5 & 43.0 & 38.5 & 15.0 & 55.9 \\
\rowcolor{blue!10}
\methodname{} (Ours) & \textbf{92.5} & \textbf{74.5} & \textbf{88.9} & 44.0 & \textbf{45.0} & 38.0 & \textbf{16.0} & \textbf{57.0} \\
\midrule
\multicolumn{9}{c}{\textit{Qwen3-8B}} \\
\midrule
GRPO & 82.5 & 71.5 & 83.7 & 38.0 & 42.0 & 38.0 & 14.0 & 52.8 \\
GSPO & 83.0 & 71.0 & 83.4 & 44.0 & 44.5 & \textbf{42.0} & 15.0 & 54.7 \\
ScalingInter & \textbf{91.0} & 73.0 & 86.2 & 41.5 & 40.0 & 38.0 & 14.0 & 54.8 \\
MATPO & 89.0 & 68.0 & 81.6 & 45.0 & 46.0 & 41.0 & \textbf{17.0} & 55.4 \\
Critique-GRPO & 90.5 & 73.5 & 86.8 & \textbf{46.0} & 47.5 & 38.0 & 14.0 & 56.6 \\
\rowcolor{blue!10}
\methodname{} (Ours) & 90.5 & \textbf{76.0} & \textbf{88.3} & 44.0 & \textbf{50.0} & 40.0 & 16.0 & \textbf{57.8} \\
\bottomrule
\end{tabular}}
\end{table*}

%% file: tables/02_math.tex
\begin{table*}[t]
\centering
\caption{Main results on math reasoning tasks. \textbf{Bold} indicates the best performance.}
\label{tab:math}
\small
\setlength{\tabcolsep}{4pt}
\begin{tabular}{l cccccc}
\toprule
\textbf{Method} & \textbf{MATH 500} & \textbf{\begin{tabular}{@{}c@{}}\textbf{Minerva} \\ \textbf{MATH}\end{tabular}} & \textbf{\begin{tabular}{@{}c@{}}\textbf{Olympiad} \\ \textbf{Bench}\end{tabular}} & \textbf{AMC23} & \textbf{AIME24} & \textbf{Avg.} \\
\midrule
\multicolumn{7}{c}{\textit{Qwen3-8B}} \\
\midrule
Qwen3-8B & 82.0 & 41.2 & 44.1 & 67.5 & 40.0 & 55.0 \\
SFT & 83.2 & 43.8 & 46.4 & 82.5 & 40.0 & 59.2 \\
GRPO & 91.0 & 52.6 & 65.6 & 82.5 & 50.0 & 68.3 \\
Critique-GRPO & 92.6 & 52.6 & 66.2 & \textbf{95.0} & 60.0 & 73.3 \\
\rowcolor{blue!10}
\methodname{} (Ours) & \textbf{94.0} & \textbf{55.6} & \textbf{68.9} & 93.1 & \textbf{65.1} & \textbf{75.3} \\
\bottomrule
\end{tabular}
\end{table*}

%% file: chapter/05_discussion.tex
\section{Discussion} \label{sec:discussion}
\subsection{Test-Time Self-Improvement}

\begin{figure*}[h]
    \centering
    \includegraphics[width=\linewidth]{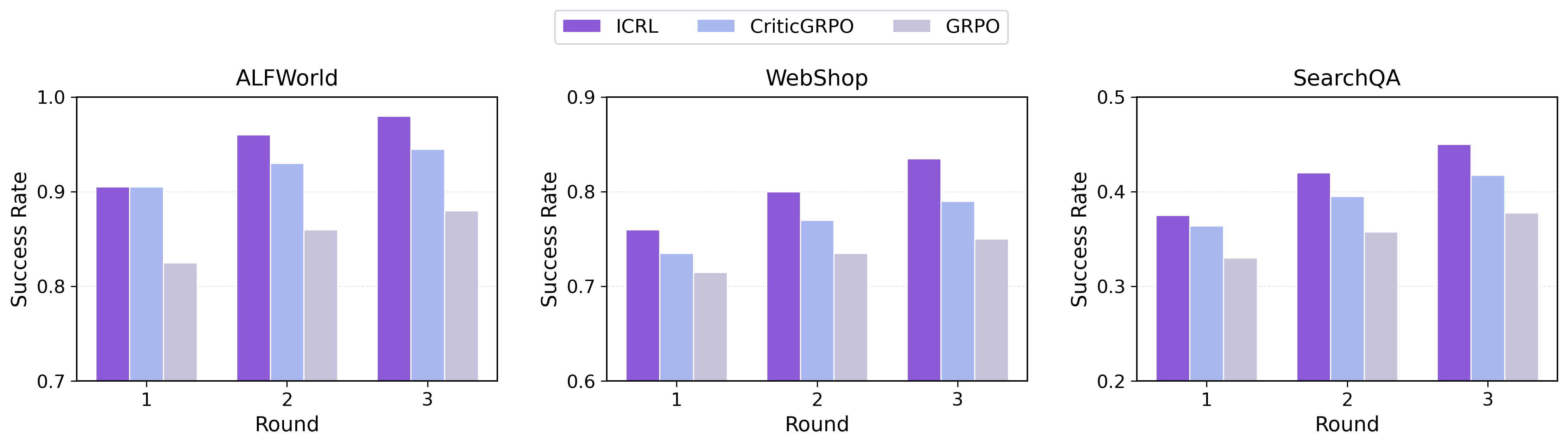}
    \vspace{-6pt}
    \caption{Test-time self-improvement performance on ALFWorld, WebShop, and SearchQA.}
    \label{fig:critic_gain_bar}
\end{figure*}

To examine whether the learned critic can further improve agents at inference time, we evaluate multi-round test-time refinement on ALFWorld, WebShop, and SearchQA. As shown in Figure~\ref{fig:critic_gain_bar}, all methods improve as the number of refinement rounds increases, but the sources of improvement differ. GRPO exhibits only gradual gains, suggesting that repeated attempts mainly provide additional sampling opportunities rather than targeted correction. Critic-GRPO improves more consistently by conditioning later attempts on critique information, but its gains remain limited across all three environments. In contrast, \methodname{} achieves both stronger first-round performance and larger improvements across refinement rounds, indicating that the learned critic provides feedback that is more useful for revising failed trajectories.

The advantage of \methodname{} is consistent across task types. On ALFWorld, \methodname{} starts near the strongest baseline and continues to improve, reaching 98\% success rate by the third round. On WebShop, it improves from 76\% to over 83.5\%, while maintaining a clear margin over both GRPO and Critic-GRPO throughout refinement. A similar trend appears on SearchQA, where \methodname{} increases from roughly 37.5\% to 45\% and consistently outperforms the critique-based baseline. These results suggest that \methodname{} does not merely benefit from additional inference-time computation; rather, its critic learns actionable diagnostic signals that help the solver identify previous errors and make more effective revisions. This is consistent with the hypothesis that joint solver-critic training can produce a self-improvement mechanism that remains useful at test time.

\subsection{Training Dynamics}

\begin{figure*}[t]
    \centering
    \includegraphics[width=\linewidth]{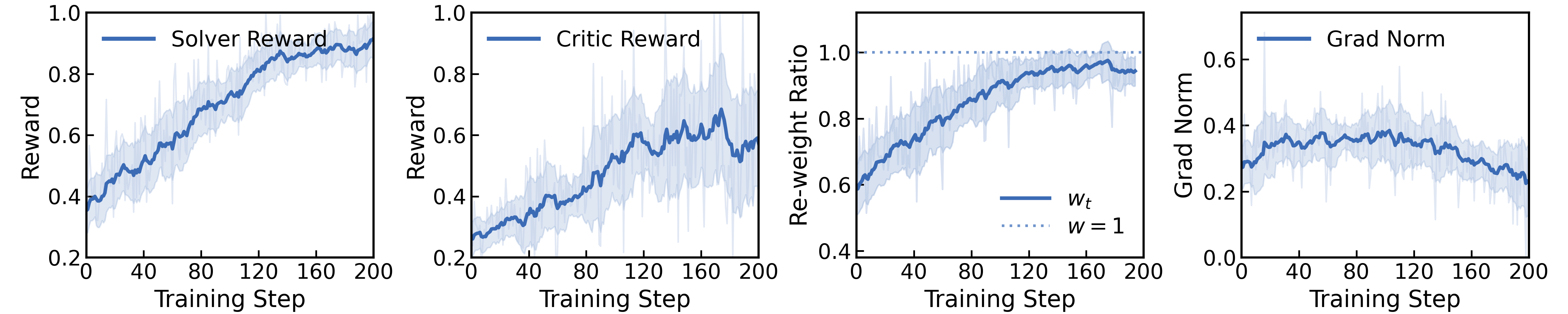}
    \caption{Training dynamics on ALFWorld. (1) Reward curve of the solver agent. (2) Reward curve of the critic agent. (3) Curve of re-weighting ratio $w_t$ (Eq.~\eqref{eq:reweight-ratio}). (4) The curve of gradient norm.}
    \label{fig:main_training_dynamics}
\end{figure*}

\textbf{Training Stability.} Stable optimization is crucial for \methodname{}, since the solver and critic are trained under a shared backbone. Figure~\ref{fig:main_training_dynamics} reports the training dynamics of \methodname{}. The solver reward steadily increases from around 40\% to above 90\%, showing that the solver improves its task performance throughout training. 
The critic reward also improves, starting from around 26\% and rising to around 60\%. This indicates that the critic learns to generate more useful critiques. It is also observed that the critic reward curve is lower and exhibits greater fluctuations, which suggests that learning to critique is more challenging and less directly tied to task success than training the solver. Meanwhile, the gradient norm remains bounded throughout training, suggesting that \methodname{} maintains stable updates and that the role-wise advantage estimation and re-weighting ratio are effective in practice.

\textbf{Re-weighting Ratio Dynamics.} The re-weighting ratio $w_t$ provides a view of how critique-guided revisions are transferred back to the critique-free solver distribution. Early in training, $w_t$ stays below the unity baseline, indicating that many successful revised actions still depend strongly on the critic context and should therefore receive a conservative update when used to train the direct solver. As training proceeds, $w_t$ increases steadily, suggesting that more tokens from critique-guided revisions become compatible with the solver's original prompt distribution. This trend supports the intended effect of distribution calibration: early training avoids blindly imitating critic-dependent behavior, while later training allows more revised trajectories to contribute once the solver begins to internalize the underlying correction pattern.

\subsection{Learning to Critique}

To investigate whether the critic learns useful feedback generation, we conduct a critic-swap evaluation by fixing the solver as \methodname{}-8B and replacing only the critic used for revision. As shown in Table~\ref{tab:critic_swap}, the learned \methodname{}-8B critic provides highly effective feedback despite being much smaller than the frozen reference critics. It achieves a 95.0\% success rate on ALFWorld, matching or surpassing larger critics, and improves WebShop performance from 76.0\% to 78.5\%. Notably, it does so with much shorter critiques, using only 57.0 tokens on ALFWorld and 93.9 tokens on WebShop, far fewer than the 20B and 32B critics. These results suggest that, in our evaluation, effective criticism is not explained by scale or verbosity alone. Instead, by optimizing the critic for downstream revision utility, \methodname{} learns solver-specific, concise, and actionable feedback that aligns with the solver's failure modes while reducing inference cost.

\begin{table}[t]
\centering
\caption{Critic-swap evaluation. The revision solver is fixed as \methodname{}-8B, and only the critic is replaced. Direct denotes no critique-based revision. Token denotes the average critique length.}
\label{tab:critic_swap}
\vspace{6pt}
\begin{tabular}{lcccccc}
\toprule
\multirow{2}{*}{\textbf{Critic}} & \multirow{2}{*}{\textbf{Deploy.}} & \multirow{2}{*}{\textbf{Size}} & \multicolumn{2}{c}{\textbf{ALFWorld}} & \multicolumn{2}{c}{\textbf{WebShop}} \\
\cmidrule(lr){4-5} \cmidrule(lr){6-7}
 & & & \textbf{SR} ($\uparrow$) & \textbf{Token} ($\downarrow$) & \textbf{SR} ($\uparrow$) & \textbf{Token} ($\downarrow$) \\
\midrule
Direct & - & - & 90.5 & - & 76.0 & - \\
\midrule
OSS-20B & Separate & 20B & \textbf{95.0} & 921.6 & 79.0 & 975.1 \\
Qwen3-32B & Separate & 32B & 94.5 & 526.6 & \textbf{80.0} & 417.7 \\
\rowcolor{blue!10}
\methodname{}-8B & Shared & 8B & \textbf{95.0} & \textbf{57.0} & 78.5 & \textbf{93.9} \\
\bottomrule
\end{tabular}
\end{table}

\subsection{Ablation Studies}

\begin{table}[t]
\centering
\caption{Ablation studies of \methodname{}. Each variant removes one component from the full method. Avg. reports the average score across the evaluated benchmarks.}
\label{tab:ablation}
\vspace{6pt}
\begin{tabular}{lccccc}
\toprule
\textbf{Variant} & \textbf{ALFWorld} & \textbf{WebShop} & \textbf{SearchQA} & \textbf{MATH} & \textbf{Avg.} \\
\midrule
\methodname{} & \textbf{90.5} & \textbf{76.0} & \textbf{37.5} & \textbf{75.3} & \textbf{69.8} \\
\quad w/o Role-wise Advantage & 88.5 & 74.5 & 36.3 & 74.1 & 68.4 \\
\quad w/o Re-weight Ratio & 88.0 & 73.0 & 35.8 & 74.2 & 67.8 \\
\bottomrule
\end{tabular}
\end{table}

As shown in Table~\ref{tab:ablation}, removing either component consistently degrades performance across all evaluated benchmarks. Without role-wise advantage normalization, the average score drops from 69.8\% to 68.4\%, indicating that solver and critic samples should not be normalized under a single shared reward scale. Since the solver is optimized for task completion while the critic is optimized for improving the solver's next attempt, role-wise normalization preserves their distinct learning signals during joint training. Removing the re-weight ratio causes a drop to 67.8\%, suggesting that the critique-conditioned response should not be optimized as a critique-free solver rollout. By re-weighting these revised trajectories, \methodname{} better aligns the training objective with the critique-free behavior distribution. Overall, the ablation results show that the gains of \methodname{} come not only from using a critic, but also from stabilizing the joint solver-critic optimization through role-aware advantage estimation and re-weight ratio adjustment.

%% file: chapter/06_related_work.tex
\section{Related Work}
\textbf{Critique and self-critique.}
Natural-language critique, feedback, and reflection have been widely used to improve language-model reasoning and agent behavior. Methods such as Self-Refine~\cite{madaan2023self-refine} and Reflexion~\cite{shinn2023reflexion} prompt models to analyze previous attempts and generate revised answers or trajectories. Other works introduce external or self-generated critics to detect mistakes, verify reasoning, or improve retrieval and generation quality~\cite{gou2023critic,Asai2023SelfRAGLT,liu2025trust_verify,chen2026learning_self_verify}. Recent studies further show that models can improve by identifying errors in their own answers or intermediate reasoning steps and revising them accordingly~\cite{weng2023large,zhang2025incentivizing,ma2025s2r}. These methods demonstrate that critique can provide effective local guidance: a failed attempt can often be repaired when the model is conditioned on appropriate feedback. However, they primarily improve critique-conditioned behavior at inference time, and do not guarantee that the model internalizes the guidance into its critique-free policy. 

\textbf{Policy Optimization.}
Policy-gradient methods have become a central tool for post-training language models and agents. Algorithms such as PPO~\cite{schulman2017proximal}, GRPO~\cite{guo2025deepseek}, and GSPO~\cite{zheng2025group} have shown strong potential in improving both reasoning models and interactive agents. Recent work studies stabilization strategies for agent RL~\cite{chen2025minimax,zheng2025stabilizing}, as well as curriculum or interaction-scaling methods for long-horizon training~\cite{xi2025agentgym-rl}. Beyond single-agent optimization, researchers have started to explore reinforcement learning for multi-agent or multi-role systems, including different agent architectures and role-specific training objectives~\cite{zhao2025stronger-mas,feng2026dr-mas,mo2025matpo}. Furthermore, expert data or expert models are incorporated into the training process~\cite{luffy,zhang2025critique,zhang2025bread}. Specifically, Critique-GRPO~\cite{zhang2025critique} leverages critic feedback for policy optimization, but still relies on a static critic model. While these methods improve agent training, they typically do not study how critique-guided revisions can be internalized into a critique-free solver.

%% file: chapter/07_conclusion.tex
\section{Conclusions}
\label{sec:conclusion}

In this work, we presented \methodname{}, a reinforcement-learning framework that jointly learns to critique and to internalize self-critique. \methodname{} converts critique-induced success into unassisted solver ability through two key mechanisms: a distribution-calibration re-weight ratio that selectively transfers critique-guided improvements compatible with the solver's critique-free distribution, and a role-wise group advantage estimation that stabilizes joint optimization of two roles. Extensive experiments across different models demonstrate its superiority over strong baselines on both agentic and mathematical reasoning tasks.
Our analyses further show that the learned critic remains useful at test time and provides concise feedback comparable to much larger frozen critics, suggesting that self-critique may be most effective when critique generation and critique internalization are learned together.

%% file: chapter/appendix.tex
\newpage
\appendix
\setcounter{figure}{0}
\setcounter{table}{0}
\renewcommand{\thefigure}{A\arabic{figure}}
\renewcommand{\thetable}{A\arabic{table}}
\renewcommand{\theHfigure}{appendix.figure.\arabic{figure}}
\renewcommand{\theHtable}{appendix.table.\arabic{table}}

\section{Analysis of Critique-Conditioned Trajectories}
\label{appendix:dist-shift}

Critique-guided trajectories are not sampled from the same behavior distribution as a critique-free solver. The initial attempt is drawn from $\pi_{\theta_{\mathrm{rollout}}}^{\mathcal S}(\tau\mid q)$, while a revised trajectory is generated under an extra critique $c$ and is therefore drawn from $\pi_{\theta_{\mathrm{rollout}}}^{\mathcal S}(\tau\mid q,c)$.
This means revised trajectories are informative, but they are distribution-shifted relative to the critique-free prompt. If they are used directly, the update targets the critique-conditioned behavior distribution rather than the intended critique-free behavior distribution, which is exactly the mismatch we want to control.

The re-weight in Eq.~\eqref{eq:reweight-ratio} is meant to transfer only the parts of a trajectory that remain plausible once the critique is removed. For a
token $y_t$ in a revised trajectory, the critique-free and critique-conditioned behavior mismatch is captured by $w_t=\pi_{\theta_{\mathrm{rollout}}}^{\mathcal S}(y_t\mid q,y_{<t})/
\pi_{\theta_{\mathrm{rollout}}}^{\mathcal S}(y_t\mid q,c,y_{<t})$. We clip this weight in practice to avoid high-variance of the gradient. The resulting update downweights tokens that depend too heavily on the critique context and keeps the solver focused on behaviors it can internalize under its original prompt distribution.

\section{Limitations}
\label{sec:limitation}
A potential limitation of \methodname{} is that it relies on the critique for revision, which may lead to longer rollout times for long-tail trajectories under synchronous RL training. In practice, this can create a bottleneck when difficult samples require multiple rounds of critique and revision before they converge. It may also reduce overall throughput if the training pipeline must wait for the slowest trajectories in each synchronous update. In our current training setup, we set the iteration round to 2 to improve training efficiency. The use of asynchronous training \cite{fu2025areal} remains unexplored and may further improve the efficiency of iterative training by decoupling these slow paths from the main optimization loop. 

\section{Training Settings}
\label{sec:config}
For all agentic environments, including ALFWorld, WebShop, and SearchQA, we use the AgentGym framework \cite{xi2025agentgym}, for math environment, we follow the same setting from Critique-GRPO~\cite{zhang2025critique}. 

\textbf{Training Settings.} We set the maximum number of agent-environment interactions to 30 turns. We use the Adam optimizer with a constant learning rate of $1 \times 10^{-6}$, a weight decay of $0.1$, and $\beta_1 = 0.9, \beta_2 = 0.98$. For each query, we sample 8 rollouts with a temperature of 1.0, a maximum context length of 16384, a maximum response length of 2048 for each turn. By default, training runs with tensor parallel size 2, sequence parallelism, dynamic batch size, and a maximum of 16384 tokens per GPU. For \methodname{} settings, the max iteration round is set to 2, and the re-weight ratio upper bound $w_\mathrm{max}$ is set to 2.

\textbf{Compute Resources.} For ALFWorld, WebShop and Math environments, Qwen3-4B experiments are trained on 2 H100 GPUs, and Qwen3-8B experiments are trained on 4 H100 GPUs. For SearchQA, we use 4 H100 GPUs and 8 H100 GPUs for 4B and 8B experiments, respectively.

\section{Environments Setting}
\label{sec:env}

\subsection{ALFWorld}
ALFWorld is a text-based household environment that aligns abstract TextWorld interaction with embodied household tasks from ALFRED \cite{ALFWorld20}.
In this environment, the agent receives a natural-language household goal and must complete it through multi-turn interaction, including navigation, object manipulation, and receptacle operations. 
ALFWorld is evaluated by task completion. The environment returns a binary outcome reward, where $r(\tau)=1$ if the final environment state satisfies the household goal and $r(\tau)=0$ otherwise.
The prompt of ALFWorld environment is as follows.

\noindent\begin{tcolorbox}[breakable, before skip=6pt, after skip=6pt, colback=white, colframe=black, title=ALFWorld, arc=3pt, boxrule=0.5pt, fonttitle=\normalfont, coltitle=white, top=2mm, bottom=2mm, left=2mm, right=2mm]
\begin{lstlisting}[style=mocov3]
You are the executor agent to complete a task in a given environment.

Environment:
ALFWorld is a household environment with rooms, receptacles, and manipulable objects.
You need to solve a household goal by navigating and interacting with objects and containers.

Task:
{{task}}

Environment action:
Available Actions:
- <action>go to [LOCATION]</action>: move to a receptacle or location.
- <action>take [OBJECT] from [RECEPTACLE]</action>: pick up an object from a receptacle.
- <action>put [OBJECT] in/on [RECEPTACLE]</action>: put an object on a receptacle.
- <action>open [RECEPTACLE]</action>: open a receptacle to reveal its contents.
- <action>close [RECEPTACLE]</action>: close a receptacle.
- <action>toggle [OBJECT] [RECEPTACLE]</action>: switch an object on or off.
- <action>clean [OBJECT] with [RECEPTACLE]</action>: clean an object using a receptacle.
- <action>heat [OBJECT] with [RECEPTACLE]</action>: heat an object using a receptacle.
- <action>cool [OBJECT] with [RECEPTACLE]</action>: cool an object using a receptacle.
- <action>inventory</action>: list objects currently being carried.
- <action>look</action>: describe the current situation.
- <action>examine [OBJECT]</action>: examine an object or receptacle in detail.

Output format:
<think>
short reason
</think>
<action>
chosen action
</action>
\end{lstlisting}
\end{tcolorbox}

\subsection{Webshop}
WebShop is an e-commerce environment in which the agent searches, navigates product pages, inspects item attributes, and purchases a product that matches a user shopping instruction \cite{yao2022webshop,xi2025agentgym}.
The task requires both information seeking and decision making because the agent must choose useful search queries and compare candidate products before purchasing.
Following the WebShop evaluation protocol \cite{yao2022webshop}, the final reward compares the purchased product $y$ with the user instruction $u$ by matching product type, attributes, options, and price:
\begin{equation}
  r
  =
  r_{\mathrm{type}}
  \cdot
  \frac{
    |U_{\mathrm{att}} \cap Y_{\mathrm{att}}|
    +
    |U_{\mathrm{opt}} \cap Y_{\mathrm{opt}}|
    +
    \mathbf{1}[y_{\mathrm{price}} \leq u_{\mathrm{price}}]
  }{
    |U_{\mathrm{att}}| + |U_{\mathrm{opt}}| + 1
  },
\end{equation}
where $U$ and $Y$ denote instruction-side constraints and purchased-product fields, and $r_{\mathrm{type}}$ penalizes mismatched product types.
An trajectory is considered successful when $r=1$, meaning the purchased product fully satisfies the instruction.
The prompt of WebShop environment is as follows.

\noindent\begin{tcolorbox}[breakable, before skip=6pt, after skip=6pt, colback=white, colframe=black, title=WebShop, arc=3pt, boxrule=0.5pt, fonttitle=\normalfont, coltitle=white, top=2mm, bottom=2mm, left=2mm, right=2mm]
\begin{lstlisting}[style=mocov3]
You are the executor agent to complete a task in a given environment.

Environment:
Webshop is an e-commerce shopping environment where you need to find and purchase product based on given shopping goal.

Task:
{{task}}

Environment action:
Every round I will give you an observation and a list of available actions, you have to respond an action based on the state and instruction.
You can use search action if search is available.
You can click one of the buttons in clickables.
If the action is not valid, perform nothing.
Keywords in search are up to you, but the value in click must be a value in the list of available actions.
Remember that your keywords in search should be carefully designed.

There are different types of pages:
- Initial Page: You can perform search actions to find products.
- Search Results Page: You can view search results, navigate through pages using click[Next >] and click[< Prev], and click on product ASIN to view details.
- Product Detail Page: You can view product details, check description and features, and if the product matches the requirements, you can purchase the product. If not, you can go back to the search results or go to initial page.

Available Actions:
- <action>search[query]</action>: search for products using the specified query (initial page only).
- <action>click[button]</action>: navigate by clicking buttons or items.

Output format:
<think>
short reason
</think>
<action>
chosen action
</action>
\end{lstlisting}
\end{tcolorbox}

\subsection{SearchQA}
SearchQA is a search-augmented question-answering environment built on multi-hop QA benchmarks, including HotpotQA, 2WikiMultiHopQA, Bamboogle, and MuSiQue \cite{HotpotQA,2WikiMultiHopQA,Bamboogle,Musique}.
The agent alternates between issuing search queries and reading retrieved information until it has enough evidence to produce a concise final answer.
Intermediate search actions are used only to gather evidence, and the trajectory is evaluated after the agent emits a final answer or reaches the interaction budget.
The environment returns a binary outcome reward based on answer correctness, with $r(\tau)=1$ when the normalized final answer matches an accepted reference answer and $r(\tau)=0$ otherwise.
The prompt of SearchQA environment is as follows.

\noindent\begin{tcolorbox}[breakable, before skip=6pt, after skip=6pt, colback=white, colframe=black, title=SearchQA, arc=3pt, boxrule=0.5pt, fonttitle=\normalfont, coltitle=white, top=2mm, bottom=2mm, left=2mm, right=2mm]
\begin{lstlisting}[style=mocov3]
You are the executor agent to complete a task in a given environment.

Environment:
SearchEnv is a websearch environment. You need to answer a question by issuing search queries and iteratively refining your search based on retrieved information.
When all necessary information is gathered, return a short and concise final answer.

Task: 
You must reason inside <think>...</think> first. If you do not have enough knowledge, issue a <search>...</search> and then STOP. Do not generate <information> or <answer> yet. Wait for external input wrapped in <information>...</information>. After receiving information, reason again in <think>. If confident, output your final answer in <answer>...</answer>. Do not output <answer> before receiving <information> unless you are fully confident. If you find no further external knowledge needed, you can directly provide the answer inside <answer> and </answer>, without detailed illustrations. For example, <answer> Beijing </answer>. Follow this process every time.
{{task}}

Environment action:
Available Actions:
- <search>query</search>: search for relevant information.
- <answer>answer</answer>: provide the final concise answer.

When giving the final answer, make it short and concise. Don't include any additional explanations or notes.
For example, if the question is "What is the capital of France?":
<answer>Paris</answer>

Output format:
<think>
short reason
</think>
<action>
chosen action
</action>
\end{lstlisting}
\end{tcolorbox}

\subsection{Math}
The math environment evaluates mathematical reasoning on MATH500, Minerva Math, OlympiadBench, AMC23, and AIME24 \cite{hendrycks2021math500,lewkowycz2022minerva,he-etal-2024-olympiadbench,numina_math_datasets}.
The agent is given a math problem and should reason step by step to arrive at the final answer. The final answer is checked against the reference answer.
The environment returns a binary outcome reward, where $r(\tau)=1$ if the final answer is equivalent to the reference solution and $r(\tau)=0$ otherwise.
The prompt of Math environment is as follows.

\noindent\begin{tcolorbox}[breakable, before skip=6pt, after skip=6pt, colback=white, colframe=black, title=Math, arc=3pt, boxrule=0.5pt, fonttitle=\normalfont, coltitle=white, top=2mm, bottom=2mm, left=2mm, right=2mm]
\begin{lstlisting}[style=mocov3]
You are a math problem solver.

Task:
{{task}}
Please reason step by step, and put your final answer within \boxed{}.

Output format:
<think>
short reason
</think>
<answer>
\boxed{answer}
</answer>
\end{lstlisting}
\end{tcolorbox}

\section{Trajectory Examples}
The following boxes show a trajectory comparison between a failed GRPO trajectory and a successful trajectory produced by our method under the same task setting.

\setcolumnwidth{0.48\textwidth,0.48\textwidth}
\begin{paracol}{2}
\noindent\begin{tcolorbox}[breakable, before skip=0pt, after skip=0pt, colback=white, colframe=black, title=GRPO, arc=3pt, boxrule=0.5pt, fonttitle=\normalfont, coltitle=white, top=2mm, bottom=2mm, left=2mm, right=2mm]
\begin{lstlisting}[style=mocov3]
Task: put a hot potato in fridge.
Response:
<|im_start|>assistant
<think>
... Let me start by going to countertop 1 to check.
</think>

<action>go to countertop 1</action><|im_end|>
<|im_start|>user
On the countertop 1, you see a dishsponge 3, a dishsponge 2, a fork 1, a spoon 1, and a tomato 1.<|im_end|>
<|im_start|>assistant
<think>
... I need to find the potato first. Let's go to countertop 2 next.
</think>

<action>go to countertop 2</action><|im_end|>
<|im_start|>user
On the countertop 2, you see a bread 2, a cup 1, and a fork 3.<|im_end|>
<|im_start|>assistant
<think>
... Let's go to countertop 3 next.
</think>

<action>go to countertop 3</action><|im_end|>
<|im_start|>user
On the countertop 3, you see a butterknife 2, a butterknife 1, a lettuce 3, a lettuce 2, a mug 1, a soapbottle 3, and a spatula 1.<|im_end|>
<|im_start|>assistant
<think>
... So perhaps go to cabinet 1, open it, and check contents. Let's try that.
</think>

<action>go to cabinet 1</action><|im_end|>
<|im_start|>user
The cabinet 1 is closed.<|im_end|>
<|im_start|>assistant
<think>... Let's open cabinet 1 to see.</think>

<action>open cabinet 1</action><|im_end|>
<|im_start|>user
You open the cabinet 1. The cabinet 1 is open. In it, you see a bowl 1.<|im_end|>
<|im_start|>assistant
<think>... Let's go to cabinet 2 and open it.</think>

<action>go to cabinet 2</action><|im_end|>
<|im_start|>user
The cabinet 2 is closed.<|im_end|>
<|im_start|>assistant
<think>
Okay, cabinet 2 is closed. Let's open it to check. If there's a potato inside, that's great. If not, proceed to the next cabinet. Alternatively, maybe check drawers. But since cabinets are common storage, perhaps continue checking cabinets. Let's open cabinet 2.
</think>

<action>open cabinet 2</action><|im_end|>
<|im_start|>user
You open the cabinet 2. The cabinet 2 is open. In it, you see nothing.<|im_end|>
<|im_start|>assistant
<think>
... Let's check drawer 1.
</think>

<action>go to drawer 1</action><|im_end|>
<|im_start|>user
The drawer 1 is closed.<|im_end|>
<|im_start|>assistant
<think>
Drawer 1 is closed. Need to open it. Let's open drawer 1 to check if there's a potato inside. If not, proceed to the next drawer or cabinet. Let's open it.
</think>

<action>open drawer 1</action><|im_end|>
<|im_start|>user
You open the drawer 1. The drawer 1 is open. In it, you see a fork 2.<|im_end|>
<|im_start|>assistant
<think>
... Let's try drawer 2.
</think>

<action>go to drawer 2</action><|im_end|>
<|im_start|>user
The drawer 2 is closed.<|im_end|>
<|im_start|>assistant
<think>
Drawer 2 is closed. Let's open it to check. If there's a potato, great. If not, continue to next drawer or check other locations. Let's open drawer 2.
</think>

<action>open drawer 2</action><|im_end|>
<|im_start|>user
You open the drawer 2. The drawer 2 is open. In it, you see a knife 1.<|im_end|>
<|im_start|>assistant
<think>
... Let's try drawer 3.
</think>

<action>go to drawer 3</action><|im_end|>
<|im_start|>user
The drawer 3 is closed.<|im_end|>
<|im_start|>assistant
<think>
Drawer 3 is closed. Let's open it to check. If there's a potato, great. If not, move to drawer 4. Let's open drawer 3.
</think>

<action>open drawer 3</action><|im_end|>
<|im_start|>user
You open the drawer 3. The drawer 3 is open. In it, you see a peppershaker 1, and a spoon 2.<|im_end|>
<|im_start|>assistant
<think>
... Let's proceed to drawer 4.
</think>

<action>go to drawer 4</action><|im_end|>
<|im_start|>user
The drawer 4 is closed.<|im_end|>
<|im_start|>assistant
<think>
... Let's open drawer 4.
</think>

<action>open drawer 4</action><|im_end|>
<|im_start|>user
You open the drawer 4. The drawer 4 is open. In it, you see nothing.<|im_end|>
<|im_start|>assistant
<think>
... Maybe the potato is in the fridge already? No, the task is to put it in the fridge. So it must be elsewhere. Let's check cabinet 3 next.
</think>

<action>go to cabinet 3</action><|im_end|>
<|im_start|>user
The cabinet 3 is closed.<|im_end|>
<|im_start|>assistant
<think>
... Let's go to cabinet 4 next.
</think>

<action>go to cabinet 4</action><|im_end|>
<|im_start|>user
On the cabinet 4, you see a soapbottle 1.<|im_end|>
\end{lstlisting}
\noindent\colorbox{red!15}{\parbox{\dimexpr\linewidth-2\fboxsep\relax}{\textbf{Reward = 0}}}
\end{tcolorbox}

\switchcolumn
\noindent\begin{tcolorbox}[breakable, before skip=0pt, after skip=0pt, colback=white, colframe=black, title=\methodname{}, arc=3pt, boxrule=0.5pt, fonttitle=\normalfont, coltitle=white, top=2mm, bottom=2mm, left=2mm, right=2mm]
\begin{lstlisting}[style=mocov3]
Task: put a hot potato in fridge.
Response:
<|im_start|>assistant
<think>
... Let's start by checking the countertops and other visible places.
</think>

<action>
look
</action><|im_end|>
<|im_start|>user
You are in the middle of a room. Looking quickly around you, you see nothing.<|im_end|>
<|im_start|>assistant
<think>
</think>

<action>
go to countertop 1
</action><|im_end|>
<|im_start|>user
On the countertop 1, you see a dishsponge 3, a dishsponge 2, a fork 1, a spoon 1, and a tomato 1.<|im_end|>
<|im_start|>assistant
<think>
</think>

<action>
go to countertop 2
</action><|im_end|>
<|im_start|>user
On the countertop 2, you see a bread 2, a cup 1, and a fork 3.<|im_end|>
<|im_start|>assistant
<think>
</think>

<action>
go to countertop 3
</action><|im_end|>
<|im_start|>user
On the countertop 3, you see a butterknife 2, a butterknife 1, a lettuce 3, a lettuce 2, a mug 1, a soapbottle 3, and a spatula 1.<|im_end|>
<|im_start|>assistant
<think>
</think>

<action>
go to stoveburner 1
</action><|im_end|>
<|im_start|>user
On the stoveburner 1, you see a pot 1.<|im_end|>
<|im_start|>assistant
<think>
</think>

<action>
go to stoveburner 2
</action><|im_end|>
<|im_start|>user
On the stoveburner 2, you see a pot 2.<|im_end|>
<|im_start|>assistant
<think>
</think>

<action>
go to stoveburner 3
</action><|im_end|>
<|im_start|>user
On the stoveburner 3, you see nothing.<|im_end|>
<|im_start|>assistant
<think>
</think>

<action>
go to stoveburner 4
</action><|im_end|>
<|im_start|>user
On the stoveburner 4, you see nothing.<|im_end|>
<|im_start|>assistant
<think>
</think>

<action>
go to microwave 1
</action><|im_end|>
<|im_start|>user
The microwave 1 is closed.<|im_end|>
<|im_start|>assistant
<think>
</think>

<action>
open microwave 1
</action><|im_end|>
<|im_start|>user
You open the microwave 1. The microwave 1 is open. In it, you see nothing.<|im_end|>
<|im_start|>assistant
<think>
</think>

<action>
go to fridge 1
</action><|im_end|>
<|im_start|>user
The fridge 1 is closed.<|im_end|>
<|im_start|>assistant
<think>
</think>

<action>
open fridge 1
</action><|im_end|>
<|im_start|>user
You open the fridge 1. The fridge 1 is open. In it, you see a apple 2, a bread 1, a egg 2, a potato 2, a potato 1, and a tomato 2.<|im_end|>
<|im_start|>assistant
<think>
</think>

<action>
inventory
</action><|im_end|>
<|im_start|>user
You are not carrying anything.<|im_end|>
<|im_start|>assistant
<think>
</think>

<action>
take potato 1 from fridge 1
</action><|im_end|>
<|im_start|>user
You pick up the potato 1 from the fridge 1.<|im_end|>
<|im_start|>assistant
<think>
</think>

<action>
go to microwave 1
</action><|im_end|>
<|im_start|>user
The microwave 1 is open. In it, you see nothing.<|im_end|>
<|im_start|>assistant
<think>
</think>

<action>
heat potato 1 with microwave 1
</action><|im_end|>
<|im_start|>user
You heat the potato 1 using the microwave 1.<|im_end|>
<|im_start|>assistant
<think>
</think>

<action>
go to fridge 1
</action><|im_end|>
<|im_start|>user
The fridge 1 is open. In it, you see a apple 2, a bread 1, a egg 2, a potato 2, and a tomato 2.<|im_end|>
<|im_start|>assistant
<think>
</think>

<action>
put potato 1 in/on fridge 1
</action><|im_end|>
<|im_start|>user
You put the potato 1 in/on the fridge 1.<|im_end|>
<|im_start|>assistant
\end{lstlisting}
\noindent\colorbox{green!15}{\parbox{\dimexpr\linewidth-2\fboxsep\relax}{\textbf{Reward = 1}}}
\end{tcolorbox}
\end{paracol}

\section{Critic Examples}
The following box shows an example of the learned critic diagnosing a failed ALFWorld trajectory and producing concise feedback for the next solver attempt.

\noindent\begin{tcolorbox}[breakable, before skip=6pt, after skip=6pt, colback=white, colframe=black, title=ALFWorld Critique, arc=3pt, boxrule=0.5pt, fonttitle=\normalfont, coltitle=white, top=2mm, bottom=2mm, left=2mm, right=2mm]
\begin{lstlisting}[style=mocov3]
<|im_start|>system
You are the critic agent. You will be given an executor trajectory that fail to complete the task. Identify the concrete failure mode and produce a short set of core tips that can help the next executor round succeed.

Environment:
{environment}

Rules:
- Diagnose the highest-value mistake.
- Focus on actionable corrections for the next executor run.
- The next round is a fresh retry from the initial observation.
- Do not output actions, XML action tags, or long narratives.
- Summarize only the reusable core tips.

Output format:
<think>
short reason
</think>
<critic>
Tip 1: ...
Tip 2: ...
Tip 3: ...
</critic><|im_end|>
<|im_start|>user
# Task
{task}
# Trajectory Summary
reward: {reward}
trajectory: {trajectory}
<|im_start|>assistant
<think>
Okay, let's look at the trajectory the executor took. The task was to cool an apple and put it in the microwave. The executor started by going to the fridge, opened it, but didn't find an apple there. Then they checked countertops and found apple 1 on countertop 3, took it, went to the sink, tried to cool it with the sink, but nothing happened. Then they cleaned it, put it in the microwave, closed it, but when trying to take it back, it didn't work. They kept looping between fridge and sink, trying to cool the apple again.

Hmm, the main issue here is that the executor didn't realize that cooling the apple requires placing it in the fridge first. The available actions might have a specific way to cool an item, like putting it in the fridge. Instead of using the sink or trying to clean it, they should have placed the apple in the fridge to cool it. Also, after cooling, they needed to take it out of the fridge and then put it in the microwave. The executor kept trying to cool it with the sink, which isn't the correct method. So the core mistakes are not using the fridge for cooling and not properly retrieving the apple after cooling.
</think>

<critic>
Tip 1: Use fridge for cooling, not sink. Cooling requires placing item in fridge, not interacting with sink.
Tip 2: After cooling, retrieve apple from fridge before placing in microwave.
Tip 3: Avoid redundant actions - don't repeatedly try to cool already cooled apple.
</critic><|im_end|>
\end{lstlisting}
\noindent\colorbox{green!15}{\parbox{\dimexpr\linewidth-2\fboxsep\relax}{\textbf{Reward = 1}}}
\end{tcolorbox}

%% file: references.bib
@article{madaan2023self-refine,
  title={Self-refine: Iterative refinement with self-feedback},
  author={Madaan, Aman and Tandon, Niket and Gupta, Prakhar and Hallinan, Skyler and Gao, Luyu and Wiegreffe, Sarah and Alon, Uri and Dziri, Nouha and Prabhumoye, Shrimai and Yang, Yiming and others},
  journal={Advances in neural information processing systems},
  volume={36},
  pages={46534--46594},
  year={2023}
}

@article{shinn2023reflexion,
  title={Reflexion: Language agents with verbal reinforcement learning},
  author={Shinn, Noah and Cassano, Federico and Gopinath, Ashwin and Narasimhan, Karthik and Yao, Shunyu},
  journal={Advances in neural information processing systems},
  volume={36},
  pages={8634--8652},
  year={2023}
}

@article{gou2023critic,
  title={Critic: Large language models can self-correct with tool-interactive critiquing},
  author={Gou, Zhibin and Shao, Zhihong and Gong, Yeyun and Shen, Yelong and Yang, Yujiu and Duan, Nan and Chen, Weizhu},
  journal={arXiv preprint arXiv:2305.11738},
  year={2023}
}

@article{Asai2023SelfRAGLT,
  title={Self-RAG: Learning to Retrieve, Generate, and Critique through Self-Reflection},
  author={Akari Asai and Zeqiu Wu and Yizhong Wang and Avirup Sil and Hannaneh Hajishirzi},
  journal={ArXiv},
  year={2023},
  volume={abs/2310.11511},
}

@article{scheurer2023training-feedback,
  title={Training language models with language feedback at scale},
  author={Scheurer, J{\'e}r{\'e}my and Campos, Jon Ander and Korbak, Tomasz and Chan, Jun Shern and Chen, Angelica and Cho, Kyunghyun and Perez, Ethan},
  journal={arXiv preprint arXiv:2303.16755},
  year={2023}
}

@inproceedings{yang2024sweagent,
  title={{SWE}-agent: Agent-Computer Interfaces Enable Automated Software Engineering},
  author={John Yang and Carlos E Jimenez and Alexander Wettig and Kilian Lieret and Shunyu Yao and Karthik R Narasimhan and Ofir Press},
  booktitle={The Thirty-eighth Annual Conference on Neural Information Processing Systems},
  year={2024},
  url={https://arxiv.org/abs/2405.15793}
}

@article{wang2024openhands,
  title={Openhands: An open platform for ai software developers as generalist agents},
  author={Wang, Xingyao and Li, Boxuan and Song, Yufan and Xu, Frank F and Tang, Xiangru and Zhuge, Mingchen and Pan, Jiayi and Song, Yueqi and Li, Bowen and Singh, Jaskirat and others},
  journal={arXiv preprint arXiv:2407.16741},
  year={2024}
}

@article{wang2024mobile,
  title={Mobile-agent: Autonomous multi-modal mobile device agent with visual perception},
  author={Wang, Junyang and Xu, Haiyang and Ye, Jiabo and Yan, Ming and Shen, Weizhou and Zhang, Ji and Huang, Fei and Sang, Jitao},
  journal={arXiv preprint arXiv:2401.16158},
  year={2024}
}

@article{qin2025ui,
  title={Ui-tars: Pioneering automated gui interaction with native agents},
  author={Qin, Yujia and Ye, Yining and Fang, Junjie and Wang, Haoming and Liang, Shihao and Tian, Shizuo and Zhang, Junda and Li, Jiahao and Li, Yunxin and Huang, Shijue and others},
  journal={arXiv preprint arXiv:2501.12326},
  year={2025}
}

@article{li2025websailor,
  title={Websailor: Navigating super-human reasoning for web agent},
  author={Li, Kuan and Zhang, Zhongwang and Yin, Huifeng and Zhang, Liwen and Ou, Litu and Wu, Jialong and Yin, Wenbiao and Li, Baixuan and Tao, Zhengwei and Wang, Xinyu and others},
  journal={arXiv preprint arXiv:2507.02592},
  year={2025}
}

@article{li2025chain,
  title={Chain-of-agents: End-to-end agent foundation models via multi-agent distillation and agentic rl},
  author={Li, Weizhen and Lin, Jianbo and Jiang, Zhuosong and Cao, Jingyi and Liu, Xinpeng and Zhang, Jiayu and Huang, Zhenqiang and Chen, Qianben and Sun, Weichen and Wang, Qiexiang and others},
  journal={arXiv preprint arXiv:2508.13167},
  year={2025}
}

@inproceedings{weng2023large,
  title={Large language models are better reasoners with self-verification},
  author={Weng, Yixuan and Zhu, Minjun and Xia, Fei and Li, Bin and He, Shizhu and Liu, Shengping and Sun, Bin and Liu, Kang and Zhao, Jun},
  booktitle={Findings of the Association for Computational Linguistics: EMNLP 2023},
  pages={2550--2575},
  year={2023}
}

@article{zhang2025incentivizing,
  title={Incentivizing LLMs to Self-Verify Their Answers},
  author={Zhang, Fuxiang and Xu, Jiacheng and Wang, Chaojie and Cui, Ce and Liu, Yang and An, Bo},
  journal={arXiv preprint arXiv:2506.01369},
  year={2025}
}

@inproceedings{ma2025s2r,
  title={S2r: Teaching llms to self-verify and self-correct via reinforcement learning},
  author={Ma, Ruotian and Wang, Peisong and Liu, Cheng and Liu, Xingyan and Chen, Jiaqi and Zhang, Bang and Zhou, Xin and Du, Nan and Li, Jia},
  booktitle={Proceedings of the 63rd Annual Meeting of the Association for Computational Linguistics (Volume 1: Long Papers)},
  pages={22632--22654},
  year={2025}
}

@article{schulman2017proximal,
  title={Proximal policy optimization algorithms},
  author={Schulman, John and Wolski, Filip and Dhariwal, Prafulla and Radford, Alec and Klimov, Oleg},
  journal={arXiv preprint arXiv:1707.06347},
  year={2017}
}

@article{guo2025deepseek,
  title={Deepseek-r1: Incentivizing reasoning capability in llms via reinforcement learning},
  author={Guo, Daya and Yang, Dejian and Zhang, Haowei and Song, Junxiao and Wang, Peiyi and Zhu, Qihao and Xu, Runxin and Zhang, Ruoyu and Ma, Shirong and Bi, Xiao and others},
  journal={arXiv preprint arXiv:2501.12948},
  year={2025}
}

@article{zheng2025group,
  title={Group sequence policy optimization},
  author={Zheng, Chujie and Liu, Shixuan and Li, Mingze and Chen, Xiong-Hui and Yu, Bowen and Gao, Chang and Dang, Kai and Liu, Yuqiong and Men, Rui and Yang, An and others},
  journal={arXiv preprint arXiv:2507.18071},
  year={2025}
}

@article{chen2025minimax,
  title={Minimax-m1: Scaling test-time compute efficiently with lightning attention},
  author={Chen, Aili and Li, Aonian and Gong, Bangwei and Jiang, Binyang and Fei, Bo and Yang, Bo and Shan, Boji and Yu, Changqing and Wang, Chao and Zhu, Cheng and others},
  journal={arXiv preprint arXiv:2506.13585},
  year={2025}
}

@article{zheng2025stabilizing,
  title={Stabilizing reinforcement learning with llms: Formulation and practices},
  author={Zheng, Chujie and Dang, Kai and Yu, Bowen and Li, Mingze and Jiang, Huiqiang and Lin, Junrong and Liu, Yuqiong and Lin, Hao and Wu, Chencan and Hu, Feng and others},
  journal={arXiv preprint arXiv:2512.01374},
  year={2025}
}

@article{liu2025trust_verify,
  title={Trust, But Verify: A Self-Verification Approach to Reinforcement Learning with Verifiable Rewards},
  author={Liu, Xiaoyuan and Liang, Tian and He, Zhiwei and Xu, Jiahao and Wang, Wenxuan and He, Pinjia and Tu, Zhaopeng and Mi, Haitao and Yu, Dong},
  journal={arXiv preprint arXiv:2505.13445},
  year={2025}
}

@article{chen2026learning_self_verify,
  title={Learning to Self-Verify Makes Language Models Better Reasoners},
  author={Chen, Yuxin and Wang, Yu and Zhang, Yi and Ye, Ziang and Cai, Zhengzhou and Shi, Yaorui and Gu, Qi and Su, Hui and Cai, Xunliang and Wang, Xiang and others},
  journal={arXiv preprint arXiv:2602.07594},
  year={2026}
}

@article{zhang2025critique,
  title={Critique-grpo: Advancing llm reasoning with natural language and numerical feedback},
  author={Zhang, Xiaoying and Zhang, Yipeng and Sun, Hao and Feng, Kaituo and Lu, Chaochao and Yang, Chao and Meng, Helen},
  journal={arXiv preprint arXiv:2506.03106},
  year={2025}
}

@article{tang2025self,
  title={Self-evolving critique abilities in large language models},
  author={Tang, Zhengyang and Li, Ziniu and Xiao, Zhenyang and Ding, Tian and Sun, Ruoyu and Wang, Benyou and Liu, Dayiheng and Huang, Fei and Liu, Tianyu and Yu, Bowen and others},
  journal={arXiv preprint arXiv:2501.05727},
  year={2025}
}

@inproceedings{ALFWorld20,
  title ={{ALFWorld: Aligning Text and Embodied
           Environments for Interactive Learning}},
  author={Mohit Shridhar and Xingdi Yuan and
          Marc-Alexandre C\^ot\'e and Yonatan Bisk and
          Adam Trischler and Matthew Hausknecht},
  booktitle = {Proceedings of the International Conference on Learning Representations (ICLR)},
  year = {2021},
  url = {https://arxiv.org/abs/2010.03768}
}

@article{yao2022webshop,
  title={Webshop: Towards scalable real-world web interaction with grounded language agents},
  author={Yao, Shunyu and Chen, Howard and Yang, John and Narasimhan, Karthik},
  journal={Advances in Neural Information Processing Systems},
  volume={35},
  pages={20744--20757},
  year={2022}
}

@inproceedings{HotpotQA,
  author       = {Zhilin Yang and
                  Peng Qi and
                  Saizheng Zhang and
                  Yoshua Bengio and
                  William W. Cohen and
                  Ruslan Salakhutdinov and
                  Christopher D. Manning},
  editor       = {Ellen Riloff and
                  David Chiang and
                  Julia Hockenmaier and
                  Jun'ichi Tsujii},
  title        = {HotpotQA: {A} Dataset for Diverse, Explainable Multi-hop Question
                  Answering},
  booktitle    = {Proceedings of the 2018 Conference on Empirical Methods in Natural
                  Language Processing, Brussels, Belgium, October 31 - November 4, 2018},
  pages        = {2369--2380},
  publisher    = {Association for Computational Linguistics},
  year         = {2018},
  url          = {https://doi.org/10.18653/v1/d18-1259},
  doi          = {10.18653/V1/D18-1259},
  bibsource    = {dblp computer science bibliography, https://dblp.org}
}

@inproceedings{2WikiMultiHopQA,
  author       = {Xanh Ho and
                  Anh{-}Khoa Duong Nguyen and
                  Saku Sugawara and
                  Akiko Aizawa},
  editor       = {Donia Scott and
                  N{\'{u}}ria Bel and
                  Chengqing Zong},
  title        = {Constructing {A} Multi-hop {QA} Dataset for Comprehensive Evaluation
                  of Reasoning Steps},
  booktitle    = {Proceedings of the 28th International Conference on Computational
                  Linguistics, {COLING} 2020, Barcelona, Spain (Online), December 8-13,
                  2020},
  pages        = {6609--6625},
  publisher    = {International Committee on Computational Linguistics},
  year         = {2020},
  url          = {https://doi.org/10.18653/v1/2020.coling-main.580},
  doi          = {10.18653/V1/2020.COLING-MAIN.580},
  bibsource    = {dblp computer science bibliography, https://dblp.org}
}

@article{Musique,
  author       = {Harsh Trivedi and
                  Niranjan Balasubramanian and
                  Tushar Khot and
                  Ashish Sabharwal},
  title        = {MuSiQue: Multihop Questions via Single-hop Question
                  Composition},
  journal      = {Trans. Assoc. Comput. Linguistics},
  volume       = {10},
  pages        = {539--554},
  year         = {2022},
  url          = {https://doi.org/10.1162/tacl\_a\_00475},
  doi          = {10.1162/TACL\_A\_00475},
  bibsource    = {dblp computer science bibliography, https://dblp.org}
}

@inproceedings{Bamboogle,
  author       = {Ofir Press and
                  Muru Zhang and
                  Sewon Min and
                  Ludwig Schmidt and
                  Noah A. Smith and
                  Mike Lewis},
  editor       = {Houda Bouamor and
                  Juan Pino and
                  Kalika Bali},
  title        = {Measuring and Narrowing the Compositionality Gap in Language Models},
  booktitle    = {Findings of the Association for Computational Linguistics: {EMNLP}
                  2023, Singapore, December 6-10, 2023},
  pages        = {5687--5711},
  publisher    = {Association for Computational Linguistics},
  year         = {2023},
  url          = {https://doi.org/10.18653/v1/2023.findings-emnlp.378},
  doi          = {10.18653/V1/2023.FINDINGS-EMNLP.378},
  bibsource    = {dblp computer science bibliography, https://dblp.org}
}

@article{zhao2025stronger-mas,
  title={Stronger-MAS: Multi-Agent Reinforcement Learning for Collaborative LLMs},
  author={Zhao, Yujie and Hu, Lanxiang and Wang, Yang and Hou, Minmin and Zhang, Hao and Ding, Ke and Zhao, Jishen},
  journal={arXiv preprint arXiv:2510.11062},
  year={2025}
}

@article{mo2025matpo,
  title={Multi-Agent Tool-Integrated Policy Optimization},
  author={Mo, Zhanfeng and Li, Xingxuan and Chen, Yuntao and Bing, Lidong},
  journal={arXiv preprint arXiv:2510.04678},
  year={2025}
}

@article{feng2026dr-mas,
  title={Dr. MAS: Stable Reinforcement Learning for Multi-Agent LLM Systems},
  author={Feng, Lang and Zheng, Longtao and He, Shuo and Zhang, Fuxiang and An, Bo},
  journal={arXiv preprint arXiv:2602.08847},
  year={2026}
}

@inproceedings{xi2025agentgym,
  title={Agentgym: Evaluating and training large language model-based agents across diverse environments},
  author={Xi, Zhiheng and Ding, Yiwen and Chen, Wenxiang and Hong, Boyang and Guo, Honglin and Wang, Junzhe and Guo, Xin and Yang, Dingwen and Liao, Chenyang and He, Wei and others},
  booktitle={Proceedings of the 63rd Annual Meeting of the Association for Computational Linguistics (Volume 1: Long Papers)},
  pages={27914--27961},
  year={2025}
}

@article{xi2025agentgym-rl,
  title={Agentgym-rl: Training llm agents for long-horizon decision making through multi-turn reinforcement learning},
  author={Xi, Zhiheng and Huang, Jixuan and Liao, Chenyang and Huang, Baodai and Guo, Honglin and Liu, Jiaqi and Zheng, Rui and Ye, Junjie and Zhang, Jiazheng and Chen, Wenxiang and others},
  journal={arXiv preprint arXiv:2509.08755},
  year={2025}
}

@misc{hendrycks2021math500,
      title={Measuring Mathematical Problem Solving With the MATH Dataset}, 
      author={Dan Hendrycks and Collin Burns and Saurav Kadavath and Akul Arora and Steven Basart and Eric Tang and Dawn Song and Jacob Steinhardt},
      year={2021},
      eprint={2103.03874},
      archivePrefix={arXiv},
      primaryClass={cs.LG},
      url={https://arxiv.org/abs/2103.03874}, 
}

@misc{lewkowycz2022minerva,
      title={Solving Quantitative Reasoning Problems with Language Models}, 
      author={Aitor Lewkowycz and Anders Andreassen and David Dohan and Ethan Dyer and Henryk Michalewski and Vinay Ramasesh and Ambrose Slone and Cem Anil and Imanol Schlag and Theo Gutman-Solo and Yuhuai Wu and Behnam Neyshabur and Guy Gur-Ari and Vedant Misra},
      year={2022},
      eprint={2206.14858},
      archivePrefix={arXiv},
      primaryClass={cs.CL},
      url={https://arxiv.org/abs/2206.14858}, 
}

@inproceedings{he-etal-2024-olympiadbench,
    title = "{O}lympiad{B}ench: A Challenging Benchmark for Promoting {AGI} with Olympiad-Level Bilingual Multimodal Scientific Problems",
    author = "He, Chaoqun  and
      Luo, Renjie  and
      Bai, Yuzhuo  and
      Hu, Shengding  and
      Thai, Zhen  and
      Shen, Junhao  and
      Hu, Jinyi  and
      Han, Xu  and
      Huang, Yujie  and
      Zhang, Yuxiang  and
      Liu, Jie  and
      Qi, Lei  and
      Liu, Zhiyuan  and
      Sun, Maosong",
    booktitle = "Proceedings of the 62nd Annual Meeting of the Association for Computational Linguistics (Volume 1: Long Papers)",
    month = aug,
    year = "2024",
    address = "Bangkok, Thailand",
    publisher = "Association for Computational Linguistics",
    url = "https://aclanthology.org/2024.acl-long.211/",
    doi = "10.18653/v1/2024.acl-long.211",
    pages = "3828--3850",
}

@misc{numina_math_datasets,
  author = {Jia Li and Edward Beeching and Lewis Tunstall and Ben Lipkin and Roman Soletskyi and Shengyi Costa Huang and Kashif Rasul and Longhui Yu and Albert Jiang and Ziju Shen and Zihan Qin and Bin Dong and Li Zhou and Yann Fleureau and Guillaume Lample and Stanislas Polu},
  title = {NuminaMath},
  year = {2024},
  publisher = {Numina},
  journal = {GitHub repository},
  howpublished = {\url{https://github.com/project-numina/aimo-progress-prize/blob/main/report/numina_dataset.pdf}}
}

@article{yang2025qwen3,
  title={Qwen3 technical report},
  author={Yang, An and Li, Anfeng and Yang, Baosong and Zhang, Beichen and Hui, Binyuan and Zheng, Bo and Yu, Bowen and Gao, Chang and Huang, Chengen and Lv, Chenxu and others},
  journal={arXiv preprint arXiv:2505.09388},
  year={2025}
}

@misc{comanici2025gemini25pushingfrontier,
      title={Gemini 2.5: Pushing the Frontier with Advanced Reasoning, Multimodality, Long Context, and Next Generation Agentic Capabilities}, 
      author={Gheorghe Comanici and Eric Bieber and Mike Schaekermann and Ice Pasupat and others},
      year={2025},
      eprint={2507.06261},
      archivePrefix={arXiv},
      primaryClass={cs.CL},
      url={https://arxiv.org/abs/2507.06261}, 
}

@misc{google2026gemini3flash,
  author       = {{Google DeepMind}},
  title        = {{Gemini 3 Flash}},
  year         = {2026},
  howpublished = {\url{https://deepmind.google/models/gemini/flash/}},
  note         = {Accessed: 2026-05-01}
}

@article{fu2025areal,
  title={Areal: A large-scale asynchronous reinforcement learning system for language reasoning},
  author={Fu, Wei and Gao, Jiaxuan and Shen, Xujie and Zhu, Chen and Mei, Zhiyu and He, Chuyi and Xu, Shusheng and Wei, Guo and Mei, Jun and Wang, Jiashu and others},
  journal={arXiv preprint arXiv:2505.24298},
  year={2025}
}

@misc{luffy,
      title={Learning to Reason under Off-Policy Guidance}, 
      author={Jianhao Yan and Yafu Li and Zican Hu and Zhi Wang and Ganqu Cui and Xiaoye Qu and Yu Cheng and Yue Zhang},
      year={2025},
      eprint={2504.14945},
      archivePrefix={arXiv},
      primaryClass={cs.LG},
      url={https://arxiv.org/abs/2504.14945}, 
}

@article{zhang2025bread,
  title={Bread: Branched rollouts from expert anchors bridge sft \& rl for reasoning},
  author={Zhang, Xuechen and Huang, Zijian and Li, Yingcong and Ni, Chenshun and Chen, Jiasi and Oymak, Samet},
  journal={arXiv preprint arXiv:2506.17211},
  year={2025}
}
